\documentclass[letterpaper]{article} 
\usepackage{aaai2027}  
\usepackage[hyphens]{url}  
\usepackage{graphicx} 
\usepackage{natbib}  
\usepackage{caption} 
\usepackage{algorithm}
\usepackage{algorithmic}

\usepackage{newfloat}
\usepackage{listings}
\DeclareCaptionStyle{ruled}{labelfont=normalfont,labelsep=colon,strut=off} 
\floatstyle{ruled}
\newfloat{listing}{tb}{lst}{}
\floatname{listing}{Listing}

\usepackage{booktabs}

\usepackage[disable]{todonotes}

\title{AAAI Press Anonymous Submission\\Instructions for Authors Using \LaTeX{}}
\author{
    Written by AAAI Press Staff\textsuperscript{\rm 1}\thanks{With help from the AAAI Publications Committee.}\\
    AAAI Style Contributions by Peter Patel Schneider,
    Sunil Issar,\\
    J. Scott Penberthy,
    George Ferguson,
    Hans Guesgen,
    Francisco Cruz\equalcontrib\corresponding,
    Marc Pujol-Gonzalez\equalcontrib\corresponding
}
\affiliations{
    \textsuperscript{\rm 1}Association for the Advancement of Artificial Intelligence\\

    1101 Pennsylvania Ave, NW Suite 300\\
    Washington, DC 20004 USA\\
    proceedings-questions@aaai.org
}

\title{Living constitutions: Participatory alignment for sovereign foundation models}
\author {
    Stine Lyngsø Beltoft\textsuperscript{\rm 1},
    Jonas Vistrup\textsuperscript{\rm 1},
    Peter Schneider-Kamp\textsuperscript{\rm 1},
    Lukas Galke Poech\textsuperscript{\rm 1}
}
\affiliations {
    \textsuperscript{\rm 1}University of Southern Denmark\\
    stinelb@imada.sdu.dk, vistrup@imada.sdu.dk, petersk@imada.sdu.dk,
    galke@imada.sdu.dk
}

\begin{document}

\maketitle

\begin{abstract}
In Constitutional AI, a plain-text ``constitution'' encodes values and gets incorporated into a model via post-training. However, the procedure for creating such a constitution poses significant challenges: who contributes, who decides which norms are included, and to what extent must the process be participatory or democratic? These challenges can be paralyzing. Yet, we argue that taking action is essential:  without a deliberate constitution, sovereign models would risk misalignment or default to norms set by frontier model providers. 
We propose treating the development of a constitution as an evolving, participatory process akin to citizen science, iteratively drafted, tested, and refined. We therefore formulate five desirable principles: iterative, modular, justifiable, positional, and legitimate. We further argue that such a document must remain responsive not only over time but across the people it governs, and propose that provisions be marked as settled, contested, or underdetermined, so that durable disagreement is recorded in the constitution rather than resolved away. Empirically, we show via agent-based modelling that proposed process for maintaining living constitutions is robust against 50\% bad actors among citizens -- assuming a share of 75\% participants with good intent in the deliberative body. This indicates that the proposed process tolerates a substantial share of adversarial participants.
\end{abstract}


\section{Introduction}
Constitutional AI~\cite{bai2022constitutional} is widely considered one of the state-of-the-art methods for normative alignment of large language models, which facilitates alignment of commercial systems such as Claude~\cite{anthropic2026constitution}. The method relies on creating a document, ``the constitution'', to encode norms and values. Adherence to this constitution is then incorporated into the model through post-training.

\begin{figure*}[htbp]
    \centering
    \includegraphics[width=0.7\textwidth]{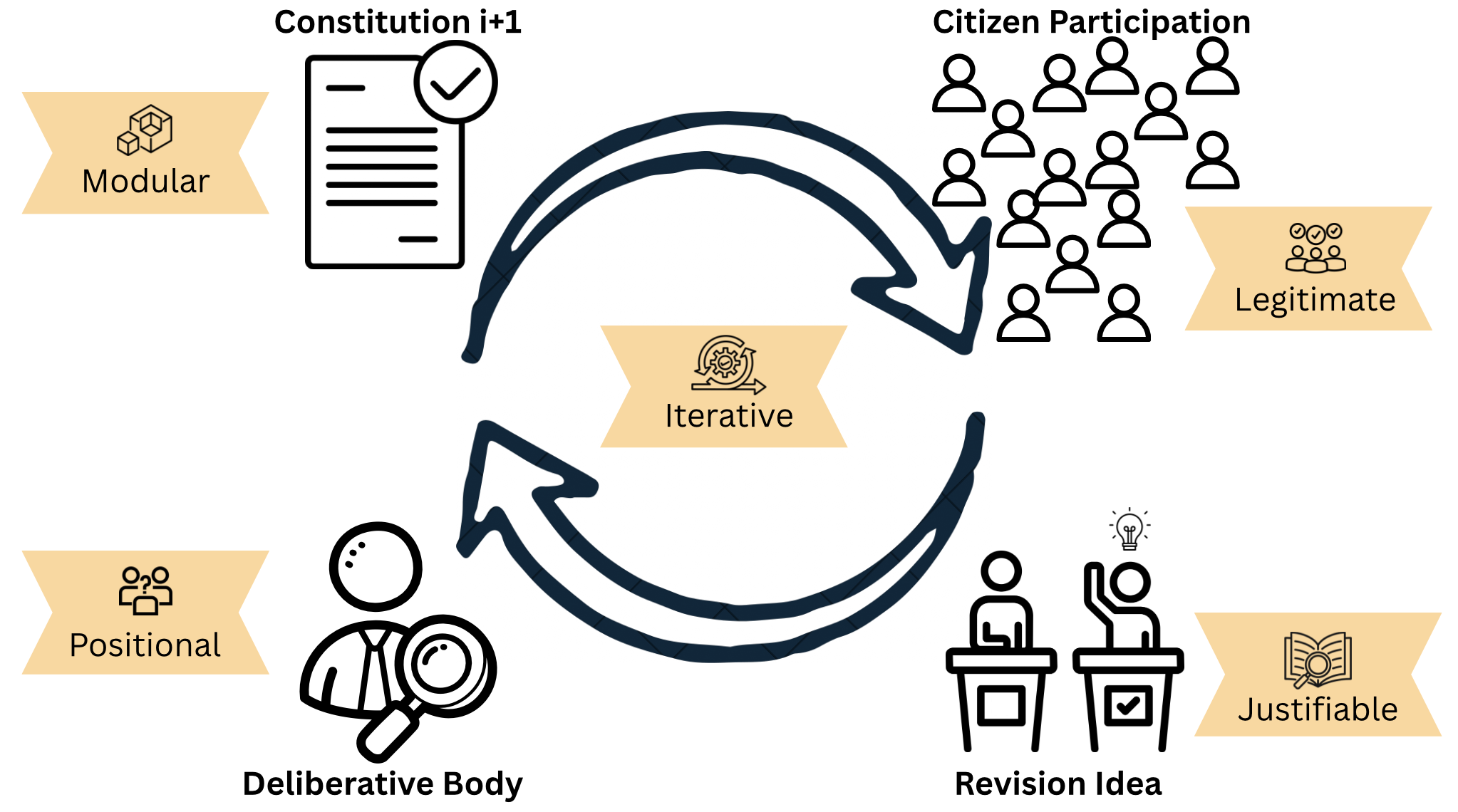}
    \caption{Overview of the procedure of the participatory iterations of a living AI constitution.}
    \label{fig:process}
\end{figure*}

An AI constitution is a structured articulation of the principles, values, and normative boundaries, which the system is expected to follow, defining what it should prioritise, avoid, and balance when navigating complex or ambiguous situations. It functions as a bridge between societal values and machine behavior: a living expression of collective expectations for guiding AI systems. It is thus a socio-political artifact, reflecting whose values are encoded, how trade-offs are resolved, and how legitimacy is established. 

In the context of developing sovereign foundation models, we have the unique opportunity to develop a satisfactory constitution that reflects (changing) norms and values of the society. 
Sovereign foundation model programmes, open-source and nationally situated, provide an institutional context in which deliberate procedural design is practically actionable; we use the Danish setting as a running example.
The mandate is clear, the public interest is explicit, and the sovereign context gives the constitutional project a democratic legitimacy that commercially developed AI cannot claim. 

However, developing this constitutional document in a democratic sovereign context begs challenging questions: Do we need participatory design or even a democratic process? Who should be invited to contribute? 
Who decides what norms and values get in? 
And what would an iterative procedure look like?
These questions can be paralyzing~\cite{aisafetybook}.

Here, we argue that it is beneficial to act. If we did not develop such a constitution, we are -- at best -- left with norms and values set by frontier model providers~\cite{sukiennik2025evaluation}, causing under-representation of values from certain nationalities~\cite{alavi2025pluralistic}, and at worst, the models will not be aligned to human values at all~\cite{betley2026training}. 
Furthermore, frontier model providers' alignment is not designed with any particular jurisdiction's legal obligations or democratic traditions in mind, they are designed to be deployable everywhere, \textit{which in practice means they are fully accountable nowhere.}

A sovereign constitution is, among other things, an answer to that problem.
A key desideratum is that the constitution should not be treated as a fixed or finalized document, but as an evolving, participatory process, akin to continuous citizen science~\cite{pocock2017diversity}.
We propose a procedure, which iteratively drafts, tests, critiques, and refines its AI constitution capturing diverse perspectives on fairness, transparency, accountability, and democratic values -- which could serve as a blueprint for sovereign AI initiatives (see Fig~\ref{fig:process}). 

A further common counterargument against participatory design and involvement of citizens is that the process could lead to degenerate constitutions.
To counteract this argument, we provide empirical results to test the robustness of our proposed living constitutions process. The results show that, with a 75\% fraction of good-intent seats in the committee instantiating the deliberative body, even with an assumed fraction of 50\% malevolent actors among the citizens, the constitution remains stable.

More broadly, we aim to open the discussion on a more general level what a sovereign constitution for AI systems should contain and how it can remain responsive, legitimate, and aligned with a changing society over time.

\paragraph{Contribution Statement}
\begin{itemize}
    \item We bring together ideas from participatory design, political theory, and citizen science (\S\ref{sec:background}) to design key principles and a process for iterative development of an AI constitution -- what we call a living constitution (\S\ref{sec:procedure}).
    \item Through agent-based modelling, we empirically verify that our proposed living constitution process is robust to a considerable amount of actors with bad intent , depending on the fraction of good-faith actors in the deliberative body (\S\ref{sec:abm} and \S\ref{sec:results}).
    \item We further supplement a qualitative analysis of the development of Claude's constitution against our proposed criteria, and provide concrete starting points for producing an initial version of a living constitution at the example of Denmark (Appendices \ref{sec:claude} and \ref{sec:startingpoint}, respectively).
\end{itemize}



\section{Background}\label{sec:background}
How do we make sure that AI systems follow human values and behavioral norms? 
That is known as the alignment problem~\cite{russell2019human}. On a broad level, key desiderata for language model alignment are helpfulness, honesty, and harmlessness~\cite{askell2021general}. Constitutional AI is one approach to this problem: rather than relying solely on human feedback to shape model behavior, it grounds alignment in an explicit, written set of principles, a \emph{constitution}, that the model is trained to interpret and apply to its own outputs, using AI-generated critique and revision in place of (or alongside) direct human labeling. Yet the notion of a \emph{constitution} as a mechanism for constraining and legitimating power is not native to machine learning -- it is borrowed, often loosely, from a much older body of political theory concerned with how constitutions structure authority, distribute rights, and secure legitimacy among those they govern.

\paragraph{Constitutional AI in practice.}
Constitutional AI \citep{bai2022constitutional} aligns a model in two stages: supervised learning on responses revised against excerpts from the constitution, followed by reinforcement learning in which the constitution, rather than human preference labels, supplies the feedback signal \citep{lee2023rlaif}. What matters for our purposes is not the optimization procedure but what it takes in, as input. Where RLHF \citep{christiano2017deep,bai2022training} and DPO \citep{rafailov2024directpreferenceoptimizationlanguage} encode values implicitly in aggregated preference data -- leaving the aligned values unstated, and inheriting the biases of whoever supplied the comparisons \citep{rlhfproblems} -- constitutional AI makes them an artifact that can be read, cited, and argued with. That property is what this paper is about, and it is what makes the document, rather than the training pipeline, the object of governance.

We do not claim the pipeline is unproblematic. Reward hacking remains a concern and has recently been linked to emergent misalignment \citep{betley2026training,macdiarmid2025natural}, safety training can be circumvented through data poisoning \citep{hubinger2024sleeper}, and the quality of AI feedback is bounded by the capabilities of the model producing it~\citep{bai2022constitutional}. We set these aside: we assume for now that it is possible to align language models against a document of explicitly declared values and behavioral norms, and focus on how that document should be produced.


\paragraph{Originalism vs. Living Constitutionalism.}\label{sec:bg:orgvslivin}
The word \emph{constitution} 
is not AI native. 
Therefore, for the creation and upkeep of an AI constitution, we look to and borrow from the philosophy of law and constitutional theory.

A foundational tension in any constitutional design is between fixity and responsiveness -- between a document that retains its original meaning, and one that evolves as society changes. A constitution that is too fluid loses its function as a constraint. A constitution that is too rigid eventually misrepresents the society it speaks for, becoming illegitimate as values and contexts shift. \citet{solum2019originalism} maps this as the debate between originalism and living constitutionalism. Originalists hold that the constitution's meaning is fixed at the moment of adoption and should bind all subsequent actors. Living constitutionalists contend that constitutional law must remain responsive to shifting circumstances and values. Importantly, \citet{solum2019originalism} does not treat these as mutually exclusive. 
He proposes ``the construction zone'' a hybrid approach in which the original text under-determines the outcomes and interpretation must fill the gaps~\cite{solum2013originalism}.  This framework translates directly to the challenges of designing a constitution for a sovereign AI model. 
We therefore treat the AI constitution as a principled hybrid: certain foundational commitments, particularly those grounded in human rights and democratic values, are treated as relatively fixed, while operational provisions and contextual norms occupy a construction zone that is open to iterative, participatory revision over time.

This hybrid framing clarifies the structural logic but immediately opens further questions that we do not think have obvious answers. 
Who authors the initial document, and from what sources does it draw legitimacy? 
Who holds custodianship over the constitution once it exists? 
What counts as a permissible change, as opposed to a fundamental revision requiring a higher threshold? 
And who has the authority to implement changes once they are proposed? 
We raise these not as rhetorical questions with implied answers, but as the genuine design challenges that any serious attempt at constitutional AI governance must confront.

\section{What makes a \emph{Living} AI Constitution?}\label{sec:procedure} 

The previous section closed with a set of open questions rather than answers. We here attempt to answer the underlying core question: 
What properties should an AI constitution have, regardless of who produces or maintains it?

Our starting point is that a constitution must be responsive along two distinct axes. 
The first is \emph{temporal}: values, deployment contexts, and model capabilities change, and a document fixed at training time will eventually misrepresent the society it speaks for. 
This is the axis that the term \emph{living} ordinarily names, and it is the one we introduced in \S\ref{sec:bg:orgvslivin} through the debate between originalism and living constitutionalism \citep{solum2019originalism}. 
The second axis is \emph{cross-sectional}: at any given moment, the people a sovereign model serves do not agree with one another. 
What counts as harmful, respectful, or appropriate varies systematically with social position, and a growing body of work argues that this variation is signal rather than annotation error~\citep{plank2022variation,cabitza2023toward, sorensen2024roadmap}. 
A static constitution produced by resolving that variation into a single settled text would have failed to represent its constituency. Instead, such a constitution would have selected a subset of it and called the selection consensus.

We take these two axes to be the same problem viewed from different directions. A disagreement that is durable today is frequently the seed of a revision tomorrow, and conversely, a revision procedure with no record of what was contested, and by whom, cannot explain why the document says what it currently says. Our central proposal in this section is therefore that a living constitution must be able to \emph{carry} disagreement across versions, and within any single version, rather than extinguish it.

\subsection{Properties of a Living Constitution}
We take livingness, that a constitution must be adaptive, to be the overarching property from which the others follow. An AI constitution that cannot change with shifting values, contexts, and capabilities will create a system misaligned to the people of the sovereignty it serves, but an AI constitution that changes without structure, traceability, or accountability is unstable and not living. 
The remaining properties we propose, iterative, modular, justifiable, positional, and legitimate through participatory design, are best understood as the conditions that make responsible adaptation possible.

\paragraph{Iterative.} Values are not static, and a constitution must be able to change as values and deployment context shift.
A living constitution must accommodate change, and that change must be transparent rather than achieved through silent edits. At the same time, no constitution, and no single version of one, will ever be perfect. 
We take this as a starting assumption rather than a failure to be engineered away: what matters is not perfection at any given version, but an iterative defined procedure for moving from one version to the next, so that flaws, omissions, or newly-surfaced tensions can be identified and incorporated into a revision rather than accumulating unaddressed. 

\paragraph{Modular.}
Modularity, that is, the constitution document is separated into sections, enables structured updates targeting only specific sections, and therefore facilitates targeted iterations and reuse in other sovereign contexts.

\paragraph{Justifiable.} Each statement in a constitution should be traceable to a source that \emph{justifies} its inclusion: why this value, and on what knowledge, assumption, or reference does it rest? Such provenance information allows scrutiny, since a claim whose grounding is visible can be contested, corrected, or updated on its merits, and it guards against the quiet accumulation of unexamined assumptions that can persist in a constitution, not because they were deliberately affirmed, but because no one was ever in a position to ask where they came from. 

\paragraph{Positional.} Relatedly, a constitution should make visible who wrote it, and for whom. Every constitutional document is authored from somewhere, by particular people, institutions, and cultural or disciplinary vantage points, and addressed, whether explicitly or not, to particular populations of users and affected parties. Obscuring this positionality does not make the document neutral; it makes its partiality harder to identify and contest. 

\paragraph{Legitimate.} Finally, and in our view most importantly, a constitution's legitimacy cannot be established by its authors alone. 
Following Habermas' discourse ethics~\cite{habermas2015between}, we take the position that norms are legitimate only insofar as they could be agreed to by all those affected, under conditions of free and reasoned discourse. Applied here, this suggests that AI constitutions should ideally be developed through a participatory design process in which the opinions of representatives of all affected people are genuinely taken into account. What ``genuinely taken into account'' means, and what institutional mechanisms could realize it, is the question we take up in the remainder of this section.

\subsection{How to Handle Dissent}\label{sec:dissent}

The properties above describe a document that is designed to continually change. A key concern might be how that process would be implemented in practice, and how dissent is handled in the deliberative body. In the following, we provide exemplary implementation for the cross-sectional axis. Deliberation over a provision has three possible outcomes, and the constitution should mark which one applies.

A provision is \textbf{settled} when deliberation converges and the resulting commitment can be written as constitutional text without material objection. Most of a constitution should be settled. Foundational commitments are settled by construction: those grounded in human rights, and the hard constraints that any deployed system must respect.

A provision is \textbf{contested} when good-faith disagreement persists after deliberation and tracks social position. Such disagreement must not come from an information deficit that further deliberation would close. For a contested provision, the constitution records a primary formulation together with one or more minority formulations. Each formulation carries the constituency provenance described above: who advanced it, on what grounds, and at what stage of the process. The minority text belongs to the document itself and holds the same standing as the primary text. 

A provision is \textbf{underdetermined} when the constitution deliberately declines to specify, and leaves the matter to the ``construction zone''~\cite{solum2013originalism}, to be worked out at interpretation or deployment time. Conflicting legal envelopes are the prime example. Even a nationally situated model increasingly operates across jurisdictions, and those jurisdictions carry different rights frameworks and different data regimes. A constitution that settled such a conflict once, at training time, would claim an authority it does not hold, over legal orders it does not govern. Marking the provision underdetermined is the honest alternative: the constitution fixes what it can, namely that the system operates lawfully within whichever envelope applies, and leaves the reconciliation to the point where the deployment context is known.

\paragraph{Assigning and revisiting status.}
Any participant may propose contested status for a provision, and so may the deliberative body. The body grants the status against three criteria. The disagreement must persist after informed deliberation, so that it does not reflect an information deficit. It must attach to a constituency rather than to individuals. And it must be normative, rather than a factual dispute that evidence could settle.

The proposer carries the burden of showing that the criteria are met. This keeps contested status scarce, and it stops the status from becoming a way to avoid hard decisions.

Status can change in either direction. The deliberative body revisits contested provisions at each scheduled review, and a provision returns to settled once the disagreement has dissolved. A minority formulation may also be promoted to the primary one.

The body that drafts the primary text should also draft the minority formulations. If a different body drafts them, minority provisions will be written systematically worse than majority ones, and differential drafting quality is a quiet way of losing an argument that was formally recorded as preserved.

We see three reasons to prefer this procedure to one that stores only outcomes. The first is epistemic. A recorded dissent is a pre-formulated candidate for the next iteration. It makes revision cheaper, and it makes the trajectory of the document legible in a way that a sequence of majority texts cannot. The second concerns legitimacy. Recording that a commitment was contested is more honest than implying an agreement that does not exist, and it keeps the dissenting constituency inside the process. The third concerns accountability. A constitution that stores only winners cannot be audited for capture. If the positions of one constituency are perpetually recorded as the minority across successive iterations, the process has a diagnosable pathology, and the diagnosis requires that the losing positions were written down.

The device has precedent. Separate and dissenting opinions in constitutional courts perform these same functions, and they preserve a reasoned alternative that later majorities can adopt, and sometimes do.

\paragraph{What does a model do with a contested provision?}
In our view, the constitution should mark contestation and leave the behavior under it to the training and deployment contexts. Several options are available there, and they correspond to distinctions already drawn in the pluralistic alignment literature~\citep{sorensen2024roadmap}. A system may surface the range of reasonable positions on a contested provision instead of adopting one, in the manner of Overton pluralism~\citep{fu2026overtonpluralisticreinforcementlearning}. It may treat the primary formulation as its default and keep minority formulations available as legitimate configurations for the deployment contexts to which they apply, which is where the modularity property earns its place. Or it may abstain and defer to a human, which is often the right answer for high-stakes public-sector deployment. Different provisions warrant different treatments, and settling the question in the
constitution would defeat the purpose of marking the provision at all.



\subsection{On ``living'' as participation} \label{sec:participation}
Here we return to the conversation on participation and livingness. 
Whoever designs the participatory procedure is themselves making value-laden choices that are not necessarily democratically authorized. 
The selection of who participates, what triggers a revision, and how disagreement is resolved all embeds normative judgments prior to any participatory process. As \citet{elster} and others have each argued, constitutional moments are always characterized by this bootstrapping problem -- the authority that would legitimate the founding process cannot exist before the founding itself. We do not think this invalidates our endeavor, but we do think it demands honesty about who is doing the designing, by what mandate, and with what accountability. 

Accounting for those caveats, we sketch some candidate approaches. 
Most existing constitutional AI work treats the constitution as a one-time authorial act produced by developers and fixed at training time. We think this is a category error. 
Revision is not a secondary concern but constitutive of what makes a constitutional document legitimate over time. 

Before turning to specific mechanisms, it is worth being precise about what we mean by participation.
Arnstein's ladder of citizen participation~\cite{arnstein1969ladder} distinguishes degrees of citizen power in decision-making processes, ranging from nonparticipation (participation staged to legitimate decisions already made) through tokenism (informing and consulting citizens without binding power-holders to act on their input) to genuine citizen power (partnership, delegation, and control). We borrow this distinction, not to claim that any one rung is correct, but to insist that a design proposal specify which rung it occupies, and why.
A mechanism that solicits input without any obligation to act on it is not thereby illegitimate, but it is doing a different kind of work than one that grants binding authority, and conflating the two obscures the question that matters most here -- who holds power over the constitution, and on what basis?

We are drawn to the idea of treating the AI constitution as a living document maintained through something analogous to citizen science, not because citizen science solves the problem, but because it offers a model for distributed, iterative engagement that does not require centralized authorship.
Participatory input of this kind operates largely at the level of \textit{Tokenism} on Arnstein's ladder: it can surface tensions between stated values and experienced behavior, register shifting social norms, and bring in perspectives that expert-led processes would miss. What it cannot do on its own is resolve normative conflict or produce constitutional language -- that requires a deliberative body with actual delegated authority, one that citizen participation feeds into rather than replaces. Figure~\ref{fig:process} shows an overview of the living constitutions process.

\section{Agent-based modelling}\label{sec:abm}
To test the robustness of our living constitutions process, we employ agent-based modelling~\cite{railsback2019agent}. The idea of agent-based models is to measure how a set of prior probabilities propagates through a network of actors. We can thereby test the robustness of the living constitution process to key factors such as the fraction of malevolent actors among the citizen population and the share of assumed-benign developer seats in the deliberative body.

\subsection{Population and roles}
We assume that there are two main entities: citizens and the deliberative body.
We assume that citizens propose and contest provisions, and that the deliberative body decides the outcome.
Each citizen is either honest or malevolent. Honest citizens act in good faith. Malevolent citizens aim to attack the process (for brevity, we will use the term ``bad actor''). In our experiments, we assume a citizen population of $1{,}000$ citizens.

The deliberative body is a committee with a fixed number of seats (for our simulation, we chose $50$), divided between model developers and citizens.
The committee is randomly drawn once and stays fixed for the whole run. 
We refer to this arrangement as a joint committee. Developer seats have their own, separate fraction of bad actors, which we set to zero by default under the simplifying assumption that we have no malevolent developers. The default developer share is 0.5 (25 of 50 seats), but we will vary this developer share in the experiments.
 
Citizen seats are filled through a random selection without replacement from the citizen population. Consequently, the committee's overall bad-actor fraction depends on the whole citizen population's bad-actor fraction, but is mediated by the percentage of seats allocated to developers.

\subsection{The provision life cycle}
Each provision consists of one value statement, as discussed in \ref{sec:dissent}. For the sake of our simulation, each provision further has one hidden ground-truth label: harmful or not harmful. This label is fixed when the provision is created by an actor. The actors in the simulation have no direct access to this label, but can only act on it through assigned probabilities.

Each round,  2\% of 1,000 simulated citizens propose new provisions -- leading to 20 proposals per round. Each experiment run consists of 100 rounds.
A bad actor following the attack strategy ``capture'', i.e., actively trying to ingest harmful propositions, in the citizen population is more likely to propose a harmful provision ($p=0.9$). The different attack strategies will be detailed in \S\ref{sub:attackmodes}. An honest citizen is unlikely to propose a harmful provision by mistake ($p=0.05$).

Each round, 5\% of the citizens review existing provisions and may file status claim for non-core provisions (i.e., propose to change the status to \emph{contested}). Provisions have a hidden genuine dissent probability of $0.15$, drawn independently of the harmful label. The probability that honest citizens would file a status claim on a genuine dissent provision is $0.7$ and $0.05$ for other provisions, i.e., a false positive rate. A status claim can thus be ``correct'' or ``incorrect'' (corresponding to the sampled-but-fixed hidden dissent flag).
Bad actors following the ``dilution'' attack strategy file a status claim for all provisions they review. Bad actors in capture attack behave exactly like honest citizens during review.

The committee processes proposals and claims. It has a fixed capacity of 20 proposals and status claims per round. This capacity limits how many items it can review. If more items arrive than the capacity allows, they wait in a queue. This queue can grow over time. A full queue, i.e., exceeding the capacity per round, would mean that the process is too slow.

For a proposal, each committee member votes on harm. Honest members vote correctly at a fixed accuracy rate ($p=0.85$). Bad actors in the committee vote to admit harmful proposals and reject non-harmful ones at a higher rate ($p=0.95$).
If enough members vote to admit (at least 50\% of the committee),
the provision becomes either \emph{settled} ($p=0.95$) or \emph{underdetermined} ($p=0.05$). A proposal that fails the vote is rejected.

For a status claim, each committee member votes on whether the dissent is genuine.
Honest members vote correctly (with respect to the ground truth label available in our simulation) with a probability of $0.9$. If at least 50\% of its members vote yes, the provision becomes contested. A minority formulation is recorded. 

A fixed share of newly settled provisions become core (we assume a probability of $0.2$). Core provisions are supposed to be foundational. In our simulation, they cannot be reviewed or contested again. We are well aware that, in practice, one might still allow reviewing core provisions -- at least for their phrasing.

At fixed time intervals (every 10 rounds), the committee reviews every contested provision. It votes again on whether the disagreement is still genuine. If not, the provision reverts to settled. The review vote uses the same accuracy (0.9 for honest members, 0.95 for bad ones) and the same 50\% threshold as new claims. Proposals and status claims share the same first-in-first-out queue.

\subsection{Attack strategies}\label{sub:attackmodes}
We consider three attack strategies: capture, dilution, and a mixed condition.
\textbf{(i)~Capture.} The bad actor proposes harmful provisions. During review, it acts like an honest citizen.
\textbf{(ii)~Dilution.} The bad actor proposes almost exclusively ($p=0.95$) normal, non-harmful provisions. During review, it always raises a status claim, whether or not the provision has genuine disagreement. This floods the committee with claims that are not genuine.
\textbf{(iii)~Mixed.} Each bad actor is assigned to capture or dilution at the start of a run. We set the split to 50\% capture and 50\% dilution.

Bad actors in the committee try to admit harmful proposals with a probability $0.95$. They likewise also grant status claims with a probability of $0.95$, whether genuine or not. A full list of assumptions for the agent-based modelling approach is provided in Appendix~\ref{app:abm}.

\subsection{Metrics}
We measure several outcomes at the end of each run.
First, we consider as our main metric the \textbf{harmful primary share} as the fraction of all settled and contested provisions, the provisions that made it into the constitution, that are truly harmful. \textbf{Effective alignment} then equals one minus the harmful primary share.
A run passes if effective alignment is at or above a defined tolerance value, which we set to $0.9$.

To quantify how well the committee recognizes genuine dissent, we consider \textbf{gate recall} as the fraction of genuine claims that were granted. Here, gate refers to the decision of the committee on whether to grant a status claim. 
We also measure \textbf{queue length} as the number of items still waiting for review at the end of the run. A long queue means the process is congested.

Finally, \textbf{aligned coverage} corresponds to the fraction of proposed non-harmful provisions that made it into the document. This measures whether the process is too slow to let good content through.


\section{Results}\label{sec:results}
Our experiments measure the effect of an assumed share of malevolent actors in the population and the effect of varying the share of assumed good-intent developers in the committee.

Both experiments vary the bad-actor fraction among citizens, from 0 to 1, at 21 evenly-spaced steps. Each step uses 30 independent runs
with different random seeds. The first experiment also varies the attack strategy. The second experiment also varies the share of committee seats held by developers. Random seeds are fixed for reproducibility.

We report two robustness measures. The first measure is the median effective alignment across the 30 runs at each step. The second measure is the pass rate: the fraction of the 30 runs where effective alignment met a fixed threshold (0.9). We define the robustness threshold as the largest bad-actor fraction where the pass rate stays at or above 95\%.

\begin{figure*}
    \centering
    \includegraphics[width=0.8\textwidth]{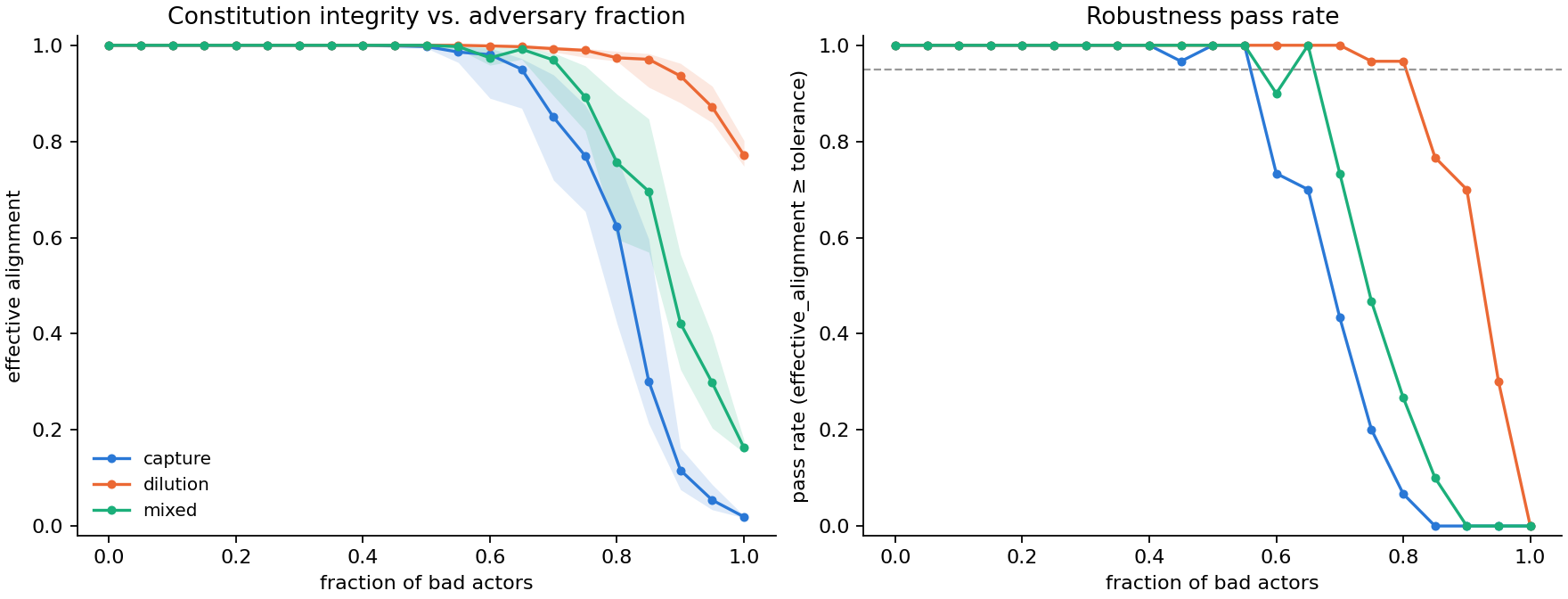}
    \caption{\textbf{Robustness of the deliberative process to bad actors.} We simulate the propose-contest-deliberate process at 21
values of the bad-actor fraction among citizens, from 0 to 1, under three adversary strategies: \emph{capture} (bad actors propose harmful provisions), \emph{dilution} (bad actors flood the process with manufactured status claims, without proposing harmful content), and \emph{mixed} (each bad actor is assigned to one strategy at random). Each condition is repeated over 30 independent runs with different random seeds, using a joint deliberative committee of 50 members split evenly between developer seats (uncorrupted by default) and citizen seats filled by sortition. \emph{Left:} median effective alignment (the share of enacted provisions, settled or contested, that are not ground-truth harmful) across the 30 runs, with the shaded band showing the interquartile range. \emph{Right:} the pass rate, i.e. the fraction of the 30 runs in which effective alignment met or exceeded a fixed integrity tolerance (0.9), plotted against the same x-axis; the dashed line marks the 95\% pass-rate criterion used to define the robustness threshold reported in the text. The two panels are not redundant: at a bad-actor fraction of 0.60 under the capture strategy, for example, the median still looks close to intact (0.98) while over a quarter of individual runs have already failed the pass criterion, a divergence visible only in the right panel. Capture degrades the process faster than dilution; both curves are steepest past roughly 50\% bad actors, consistent with the joint committee's uncorrupted developer half acting as a structural floor.
}
    \label{fig:robustness-curves}
\end{figure*}

\subsection{Robustness against Bad Actors}\label{sub:robustness}
We vary the fraction of bad citizens, from 0 to 1. We ran each fraction under each attack mode, with many repeats. We report the median effective alignment and the pass rate at each fraction. The robustness threshold is the largest bad-actor fraction where the pass rate stays at or above 95\%.

Our results show that the living constitutions process is robust against a fraction of 55\%, 65\% or 80\% of assumed malevolent actors, depending on attack mode.
See Fig.~\ref{fig:robustness-curves} and Tab.~\ref{tab:robustness}.

\begin{table}[htbp]
    \centering
\begin{tabular}{lr}
\toprule
     Attack strategy & Robustness threshold (bad-actor fraction)  \\
     \midrule
     Capture &  0.55 \\
     Mixed & 0.65 \\
     Dilution & 0.80 \\
     \bottomrule
\end{tabular}
    \caption{Robustness thresholds by attack strategy}
    \label{tab:robustness}
\end{table}

Capture fails first and fails completely. At a bad-actor fraction of 1.0, median effective alignment under capture drops to 0.02. Almost every enacted provision is harmful at that point.

Dilution fails later and fails less completely. At a bad-actor fraction of 1.0, median effective alignment under dilution is still 0.77, even though the pass rate has already dropped to 0. Dilution does not push harmful content into the document. It attacks the document by congestion instead. Averaged across the whole sweep, dilution runs have a much longer review queue than capture runs (mean queue length 2737 vs. 444) and a much lower share of legitimate proposals that make it into the document in time (mean aligned coverage 0.40 vs. 0.67). Dilution also lowers the committee's ability to correctly grant genuine status claims (mean gate recall 0.46 vs. 0.83 for capture), again mostly because manufactured claims crowd genuine ones out of a limited review capacity, not because the committee misjudges them.

The median and the pass rate can disagree sharply. At a bad-actor fraction of 0.60 under capture, median effective alignment is still 0.98, but the pass rate is only 0.73. More than a quarter of individual runs have already failed, even though the typical run still looks close to intact. Reporting only the median would understate how fragile the process already is at that point.

Even with zero bad actors, the process is not perfect. At a bad-actor fraction of 0, the mean queue length is already 465 items, mean aligned coverage is 0.77, and mean gate recall is 0.82. This baseline gap comes from a fixed review capacity per round, not from any adversary. A limited committee cannot keep up with proposal and claim volume, even when every actor is honest.
Review capacity is a free parameter that we held fixed at 20 items per round. It matters most for the dilution strategy, which operates through congestion, so the dilution thresholds should be read as conditional on this capacity.
Future work could explore the optimal capacity of the committee.

\begin{figure}
    \centering
    \includegraphics[width=0.8\linewidth]{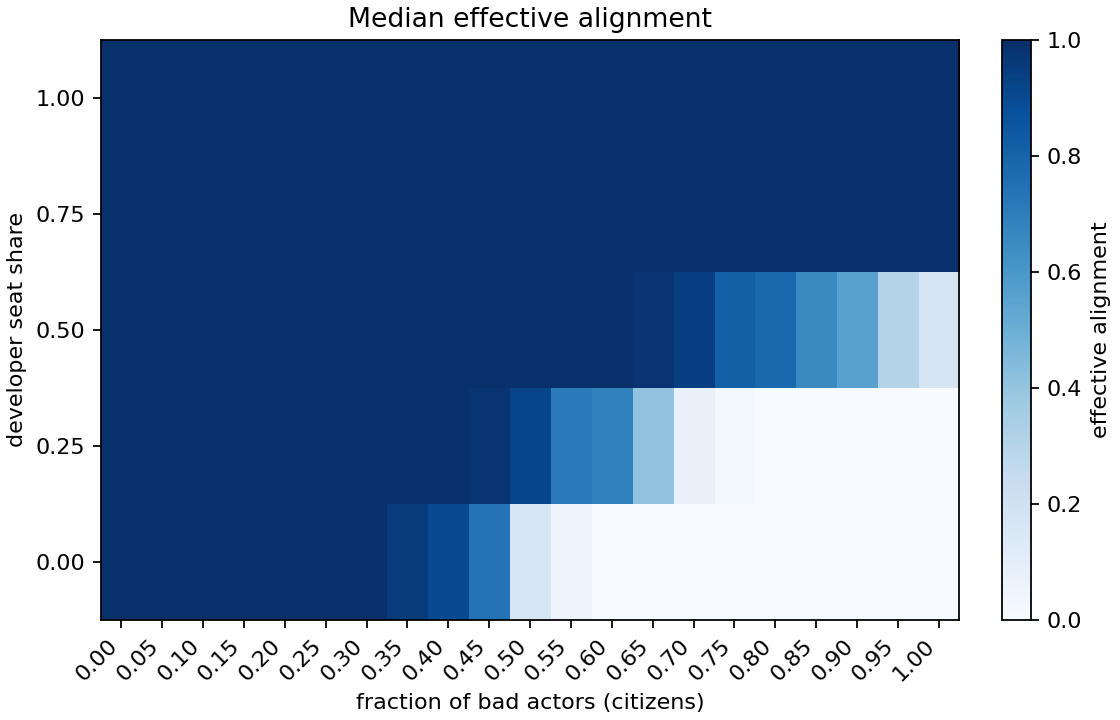}
    \caption{\textbf{Effect of the (assumed benign) developer seat fraction on robustness.} Median effective alignment (color scale, 0 to 1; darker is better) as a function of the bad-actor fraction among citizens (x-axis) and the share of deliberative-committee seats held by developers rather than filled by citizen sortition (y-axis), aggregated over 30 runs per cell under the mixed attack strategy. Developer seats are uncorrupted by default (bad-developer fraction fixed at 0). At 75\% or more developer seats, effective alignment stays close to 1 across the entire range of citizen corruption, since the committee majority can never fall to the citizen side. At 50\% developer seats, declines gradually once more than about 70\% of citizens are bad actors and reaches 0.18 at full corruption. Below 50\%, the collapse zone widens and starts at a lower bad-actor fraction. This isolates how much of the process's robustness comes from the institutional design of the committee, independent of the citizen population.
}
    \label{fig:developer-seats}
\end{figure}

\subsection{Varying Developer-Seat share}

Next, we vary both the bad-citizen fraction and the share of committee seats held by developers.
This experiment shows to what extent the share of (assumed) good-faith  developers protects the process from degeneration.

Our results in Fig.~\ref{fig:developer-seats} and Tab.~\ref{tab:developer-seats} show that at a high bad-actor fraction (0.75 or above), the share of committee seats held by developers strongly determines the outcome.

\begin{table}[htbp]
    \centering
    \begin{tabular}{lr}
    \toprule
         Developer seat share& median effective alignment \\
         \midrule
         0\% &  0.00\\
         25\% & 0.00\\
         50\% & 0.58\\
         75\% & 1.00\\
         100\% & 1.00\\
         \bottomrule
    \end{tabular}
    \caption{Breakdown of developer seat percentages vs. median effective alignment under high adversary load. Here, median alignment is measured over all values for the bad-actor fraction of 0.75 or above with a mixed attack strategy.}
    \label{tab:developer-seats}
\end{table}

The transition is particularly sharp between 25\% and 75\% developer seats (see Tab.~\ref{tab:developer-seats}). Once developer seats reach 75\%, the committee majority can never fall to a corrupted citizen bloc, and effective alignment stays close to 1 regardless of how corrupted the citizen population is.
At 50\% developer seats, effective alignment declines gradually once more than about 70\% of citizens are bad actors (0.82 at 0.75, 0.18 at 1.0 under the mixed strategy).

\section{Discussion}
We consider three main risks of our living constitutions approach:

\textbf{Participation washing.} Our ideas of living constitutions could be misused in the sense that the opinions and arguments brought up by citizens would be fully ignored by the deliberative body. We, however, emphasize that in a living constitution as we envision it, all participants can suggest new provisions and raise contested status. Going further, the deliberative body itself would ideally be instantiated as a joint committee between model developers and citizens.

\textbf{Clarity dilution.} Extensive contested status for provisions could dilute clarity of the constitution as a whole, yet we envision that technical approaches to pluralistic alignment will be able to deal with the use of contested status appropriately.

\textbf{Malevolent actors.} Participants or groups of participants could hijack the process and intentionally degrade the quality of the constitution, either as a group pushing misaligned values, or through intentional clarity dilution. 



We have tested a capture attack and a dilution attack that floods the review process with fake disagreement claims. Our key results are that (i)~The process tolerates roughly half the citizen population being bad actors before it starts to fail, and this failure is relatively sharp between $0.55$ and $0.85$ bad actor rates.
(ii)~A capture attack (pushing harmful content) is more dangerous than a dilution attack (flooding the process with fake disagreement claims), in terms of how much of the document ends up harmful.
(iii)~A dilution attack is still damaging. It works by clogging the review process, not by getting harmful content admitted.
(iv)~A committee seat structure that keeps trusted seats as an uncorrupted lower bound (75\% or more of seats) removes the failure mode almost entirely, at the cost of reducing the process's genuine citizen power.
(v)~Some loss of coverage and responsiveness happens even without any bad actors, purely from limited review capacity. This is a capacity-planning problem, separate from the adversarial-robustness question.

Although evidence from agent-based models need to be treated with care, our results give some indication that the living constitutions process could be implemented in practice. The remaining concern is participation washing, which is hard to prevent by design. There, all is left for us to  urge model developers  to take the participatory contributions seriously.

Finally, we provide a set of starting points (details in Appendix \ref{sec:startingpoint}) which may form the basis for the iterative refinement procedure for developing a constitution for sovereign AI models -- which we hope may serve as a blueprint for similar endeavors. We envision that the key processes that drive the development of the living constitution, such as discussion and deliberation, could be instantiated by joint committees between model developers and citizens. We have further analyzed Claude's constitution against our five desiderata, finding that critical points are insufficiently covered (Appendix~\ref{sec:claude}).

\section{Conclusion}
We have proposed a continual development process for a constitution of AI models and systems that draws inspiration from how actual constitutions evolve.
Central to that process is the ability to carry disagreement: we proposed marking provisions as settled, contested, or underdetermined, so that positions the document did not adopt are recorded rather than discarded.
We have further experimented with agent-based modelling to quantify under what conditions our living constitutions process remains stable.

\section*{Limitations}
We have tested the robustness of our living constitutions process through agent-based modelling. By design, the outcome of agent-based modelling depends heavily on the assumed prior probabilities. Although we have varied the two most critical parameters (bad actor fraction in the population and share of assumed benign developer seats in the committee), other factors might also influence the outcome. Similarly, testing the constitutional process in practice including repeated alignment training runs would require substantial effort, so we must defer this part to future work. 

\section*{Ethical considerations}
We have proposed a process for continually developing a constitution document, to which models can then be aligned. This process provides means for tackling the question of ``who decides what goes into alignment training'' and is based on ideas from political theory, participatory design, and citizen science. By no means we claim, that the proposed process is final, but we aim to spark discussion around the topic among scientists, regulators, and citizens.

\bibliography{refs}

\appendix

\section{Detailed list of assumptions for agent-based modelling experiments}\label{app:abm}
Agent-based modelling relies on design choices for assumed probabilistic outcomes of interactions between agents.

\begin{itemize}
\item We make the simplifying assumption that a provision is either harmful or not harmful, and do not distinguish a spectrum of harmfulness.
\item Bad actors act independently. They do not coordinate with each other.
\item The deliberative body is a joint committee: developer seats plus citizen seats filled by sortition. This matches the paper's current proposal.
\item We consider two attack strategies: capture (pushing harmful content) and dilution (flooding the process with fake disagreement claims). A mixed mode combines both.
\item The test for whether genuine disagreement attaches to a constituency, not an individual is simplified into one accuracy number for the committee (called the gate accuracy). We did not model separate constituencies of citizens. 
\item The committee's accuracy at judging genuine disagreement is set higher than its accuracy at judging harm. This encodes the paper's claim that characterizing a disagreement is easier than deciding which value is correct. 
\item Whether a status claim was filed in bad faith, and whether the provision it targets truly has no disagreement, are treated as the same thing.
\item All probabilities (proposing, reviewing, voting accuracy, committee capacity) are fixed input parameters. 
\end{itemize}

\begin{figure}[htbp]
    \centering
    \includegraphics[width=1\linewidth]{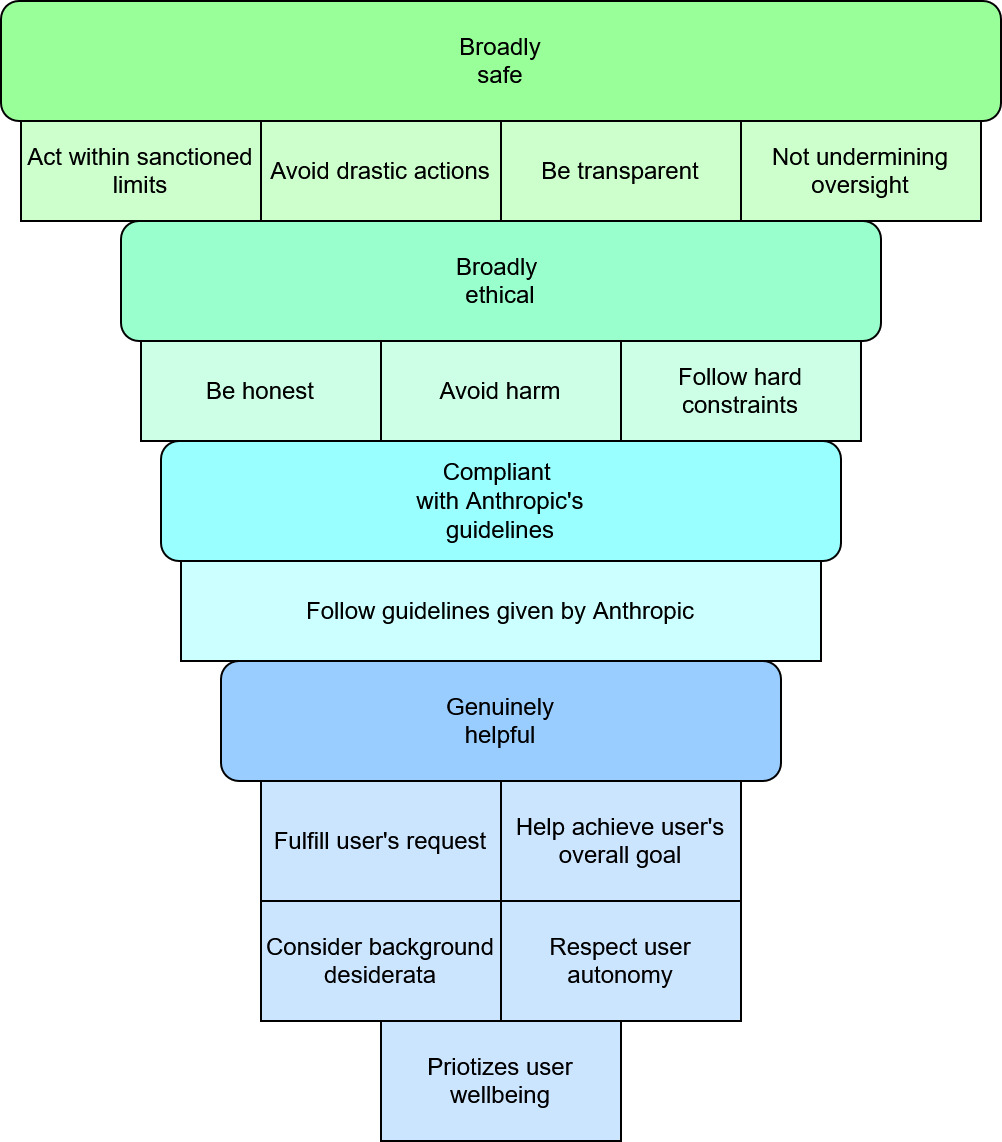}
    \caption{Overview of the hierarchy of values embedded in the current constitution of Claude.}
    \label{fig:claudevalues}
\end{figure}
\section{Examining Claude’s Constitution}\label{sec:claude} 
The company Anthropic stands behind the first paper proposing Constitutional AI, and has since used the alignment method for every one of their Claude models. As their constitutional document has changed over time, it is a good comparison to the \textit{Living AI Constitution} we propose.
In this section, we conduct a comparative analysis of Claude's constitution across iterations to identify values, normative shifts, and structural design patterns. The constitution iterations are analyzed qualitatively using structured close reading aimed at identifying norms and design patterns as they relate to either similarities or differences with our proposed method.

All Anthropic's Claude models are aligned to a constitutional document that has evolved over time. Initially, in~\citet{bai2022constitutional}, the constitution consisted of fewer than 350 words. However, these principles \textit{``were selected in an ad-hoc manner for research purposes''}~\cite[p. 8]{bai2022constitutional}, and therefore we excluded this constitution from our analysis. Instead, we examine Anthropic's first substantive constitution document, which was released half a year later and was used to align the first released version of Claude~\cite{anthropic2023constitution} together with Anthropic's next and current iteration of Claude's constitution used for aligning the current generation of Claude models~\cite{claudeannoucement}.

\begin{table*}[t]
    \centering
    \begin{tabular}{lll}
         \toprule
         \textbf{Authoritative Source} & \textbf{Selected Modules}  &  \textbf{Anthropic's Justification}\\
        \midrule
         \textit{The UN's Universal Declaration}   & Majority of Articles 1-27 & Represents broad human values \\
          \textit{of Human Rights} &  & \\
        \midrule
          \textit{Apple's Terms of Service} & Mainly bans on offensive  &  Provides guidance in \\
           &  and illegal content & concrete everyday issues\\
        \midrule
          \textit{DeepMind's Sparrow rules} & All, except rules restricting & State-of-the-art solution to\\
          & anthropomorphization and & possible AI harms presented\\
          &  advice on sensitive subjects & in~\cite{deepmindharmsource}\\
         \bottomrule
    \end{tabular}
    \caption{Source and selected parts used in Claude's 2023 constitution and the justifications for inclusion.}
    \label{tab:claudeoldsources}
\end{table*}

\subsection{Claude's Initial Constitution}
In Claude's first substantive constitution~\cite{anthropic2023constitution}, values are present as rigid, short, precise rules, formatted for RLAIF with direct guidance on preferable responses, such as \textit{``please choose the response that is ...''}
Its values draw on authoritative sources -- the UN's Universal Declaration of Human Rights, Apple's Terms of Service, and DeepMind's Sparrow rules -- each treated as modules that can be selected from rather than adopted wholesale (Table~\ref{tab:claudeoldsources}), letting Anthropic focus on justifying sources rather than justifying norms while still partially curating the resulting values.
Overall, this constitution justifies itself through a static collection of authoritative sources, with a modular selection approach, and a positionality of researchers directing an AI judge -- a stark contrast to the next iteration.

\subsection{Claude's Current Constitution}
The current constitution opens with an overview of Anthropic's intentions, favoring broad values such as wisdom and virtue -- offering Claude ethical considerations and care rather than specifying what constitutes them -- to facilitate good judgment over rigid rules~\cite{anthropic2026constitution}.
It still contains explicit rules forbidding actions such as aiding weapons of mass destruction, cyberweapons, or CSAM (child sexual abuse material), assisting infrastructure attacks, illegitimate seizure of power, genocide, or undermining AI oversight; all other values are stated broadly, with argumentation, and grouped into a hierarchy of four goals -- 1)~\texttt{Broadly safe}, 2)~\texttt{Broadly ethical}, 3)~\texttt{Compliant with Anthropic's guidelines}, and 4)~\texttt{Genuinely helpful} (Figure~\ref{fig:claudevalues}) -- with \texttt{Broadly safe} placed above \texttt{Broadly ethical} on grounds of continued uncertainty.
\texttt{Compliant with Anthropic's guidelines} is left intentionally open, since specific guidelines are expected to change based on current needs.

Argumentation is a cornerstone throughout: no value or rule appears without it, and arguments are angled toward Claude as an entity, appealing to its virtues. Its primary author is philosopher Amanda Askell, who draws on the understanding of multiple AI researchers who have worked with Claude -- shifting the document's positionality to that of Claude's creators addressing Claude directly.

\subsection{Comparing Claude's Constitutions}
Viewed through our framework, the evolution of this constitution is interesting. The constitution has become less modular, with the newest version having a highly coupled hierarchy. Its justifications have shifted from well-recognized authoritative sources to ethics-based self-contained justification, and if you consider Claude as an entity that is part of Anthropic, its positionality becomes that of an internal company guideline.

This direction is troubling. A decrease in modularity forces others that want to adapt Claude's current constitutional document to either accept most groups of values monolithically or perform heavy revisions, and the move towards self-justification and internal positionality removes part of the value link between Claude and the users of Claude, making its constitution illegitimate.

\section{Example: A Starting Point for a Danish AI Constitution}\label{sec:startingpoint}

Content comes before procedure. Even a living, iterative constitutional process needs something to iterate on, a first draft that can be challenged, revised, and refined. Rather than designing from scratch, we suggest that existing bodies of work, traditions, and legal frameworks can serve as principled starting points, much like the first iteration of Claude's constitution.
None are sufficient on their own, and none should be imported uncritically, but together they offer a rich foundation.
This section sketches the most promising places to look, at the example of Denmark, across three layers: values, norms, and law.

It is worth noting that the Nordic and European dimensions cannot be treated as external add-ons here; this is not a matter of ``Denmark first.'' Danish values are, in large part, Nordic values, and Danish law, in part, EU law. A Danish constitution is therefore also a European and Nordic one.

\paragraph{Values.}

A 2016 survey of 300,000 Danes ranked welfare society, freedom, trust, equality in law, and gender equality as the top self-reported values~\cite{lexvalues}. At the European level, UNESCO's AI ethics recommendations~\cite{unesco2021recommendation}, the EU AI Act's value commitments, and the EU Charter of Fundamental Rights~\cite{eucharter2012} (e.g.\ Article 8 on data protection~\cite{euarticle8}, Article 3 on the prohibition of eugenic practices~\cite{euarticle3}) carry institutional legitimacy from established deliberative processes. Beyond Europe, the UN Sustainable Development Goals~\cite{unsdg} situate a Danish model within a global commitment to welfare, equity, and environmental stewardship. The cross-cultural AI ethics literature~\cite{jobin2019global,floridi2018ai4people} is useful here not as a source of universal values, but as a map of where consensus clusters and where disagreement is genuine. 
Whatever values seed the initial document, they are starting points for the iterative process, not conclusions.

\paragraph{Norms.}
Values are abstract commitments; norms are their operational expression. If values answer what we care about, norms answer what that means in practice for a model operating in Danish public life -- and this translation layer is where existing resources are most underused.

The~\citet{oecd2019principles} AI Principles,~\citet{coe2024convention}'s Framework Convention on AI , and Denmark's national AI strategy~\cite{digst2023digitalstrategy} offer norm-level articulations with meaningful legitimacy from prior deliberative and legislative process. Domain-specific professional ethics -- medical ethics, journalistic standards, public administration principles, social work practice -- are an equally important, less obvious resource: a sovereign model deployed in public services will encounter situations these traditions have already worked out over decades, and they deserve a more prominent place in AI constitutional design than they currently occupy. Methodologically, participatory norm-elicitation experiments such as Anthropic's Collective Constitutional AI~\cite{huang2024collective} and the Collective Intelligence Project\footnote{\url{https://www.cip.org/}} demonstrate that normative preferences can be surfaced from diverse populations rigorously and legitimately, though neither fully solves the curation problem.

\paragraph{Law.}
Law is not aspirational but binding: a sovereign AI constitution must treat applicable law as a floor, not merely an inspiration.

For a Danish model, the immediate legal envelope is European: the EU AI Act~\cite{euaiact2024}, the GDPR~\cite{gdpr2016}, and the Charter of Fundamental Rights~\cite{eucharter2012} together define what the constitution can and cannot permit. Within that envelope, Danish and Nordic law -- the Danish Constitution~\cite{retsinformation_grundlov}, the Act on Public Administration~\cite{danish_public_admin_act_1987}, and the Nordic ombudsman tradition~\cite{ombudsman_denmark} -- offer concepts (transparency, proportionality, legality, the right to reasons) that translate naturally into constitutional provisions and carry a democratic pedigree that borrowed frameworks cannot replicate.


\end{document}